\documentclass[conference]{IEEEtran}

\input epsf     

\usepackage[dvips]{graphicx}
\usepackage{amsmath}
\usepackage{enumerate}
\usepackage{amsfonts}
\usepackage{amssymb}
\usepackage{amscd}
\usepackage{hhline}
\usepackage{epsfig}
\usepackage{color}
\usepackage{url}

\usepackage{makeidx}

\title{Multi-Spectral Facial Biometrics in Access Control}

\author{\IEEEauthorblockN{K. Lai, S. Samoil, and S.N.Yanushkevich}
\IEEEauthorblockA{Department of Electrical and Computer Engineering\\
Biometric Technologies Laboratory, University of Calgary\\ 
2500 University Drive NW, Calgary, Alberta, T2N 1N4 Canada\\
Email: \{kelai,ssamoil,syanshk\}@ucalgary.ca}
}

\begin{document}

 \maketitle
\IEEEoverridecommandlockouts
\IEEEpubid{\begin{minipage}{\textwidth}\ \\[55pt]
		\footnotesize{{\fontfamily{ptm}\selectfont Digital Object Identifier 10.1109/CIBIM.2014.7015450 \\ 978-1-4799-4533-7/14/\$31.00 \copyright 2014 IEEE}}
\end{minipage}}

\begin{abstract}
This study demonstrates how facial biometrics, acquired using multi-spectral sensors, such as RGB, depth, and infrared, assist the data accumulation in the process of authorizing users of automated and semi-automated access systems. This data serves the purposes of person authentication, as well as facial temperature estimation. We utilize depth data taken using an inexpensive RGB-D sensor to find the head pose of a subject. This allows the selection of video frames containing a frontal-view head pose  for face recognition and face temperature reading.  Usage of the frontal-view frames improves the efficiency of face recognition while the corresponding synchronized IR video frames allow for more efficient temperature estimation for facial regions of interest.  In addition, this study reports emerging applications of biometrics in biomedical and health care solutions. Including surveys of recent pilot projects, involving new sensors of biometric data and new applications of human physiological and behavioral biometrics. It also shows the new and promising horizons of using biometrics in natural and contactless control interfaces for surgical control, rehabilitation and accessibility.
\end{abstract}

\section{Introduction}

New concepts, such as the Smart border \cite{[kn:Bigo-CEPS-2012]}, are being developed that use the vast spectrum of sensors for constant monitoring of travelers, their behavior indicators, and other measurements such as temperature that are important for the control of pandemics and epidemics at the borders. For example, initial information can be provided by ID, on-line biometrics and surveillance data, personal information from various databases, and interview data \cite{IATA-Checkpoint-future}.  

Future generations of biometric-based access control systems may likely utilize multiple resources to perform monitoring and authentication, not
only for access granting decision-making but also for situational awareness and risk management.  Among these resources, there are invariant representations of individuals using multi-spectral sensors. Recently, this spectrum has been extended due to advances in technology. For example, new devices such as RGB-Depth (RGB-D) sensors (the Microsoft Kinect is an example) and time-of-flight sensors are now available. The effect of using RGB-D sensors on the performance of facial biometrics in e-border verification procedures is studied in this paper.

  \section{Prototyping the biometric-based situational awareness and risk management}
 
 Various intelligent technologies for access control, such as border control,  are currently being developed, as reviewed in \cite{[kn:Frontex-IDCHECK]}. They include decision-making and human-machine interaction,  identity verification and biometrics, risk assessment of behavioral biometrics, travel documents, document inspection systems, fraud detection, and interview supported machines. 

This support may be accomplished by effective utilization of the automated access control topology, advanced biometric technologies, and  human-machine interaction techniques. The latter makes use of  the experience coming from well-known dialogue systems design such as, in particular, SmartKom \cite{[kn:Herzog-Reithiger]} that possesses sensor specific input processing, modality-specific analysis and fusion, and interaction management. In particular, the interview-support system for border control technologies was prototyped in the form of the AVATAR kiosk \cite{[Nunamaker-AVATAR-2011]}.  Partial prototypes of facial biometric components in biometric-based decision-making and interview-supporting systems are reported in \cite{[kn:Yanushkevich-Paris],[kn:Yanushkevich-2008],[kn:Yanushkevich-2010],[poursaberi]}.  

 \begin{figure*}[!ht!] 
			\hspace{25mm}
      \includegraphics[scale=0.5]{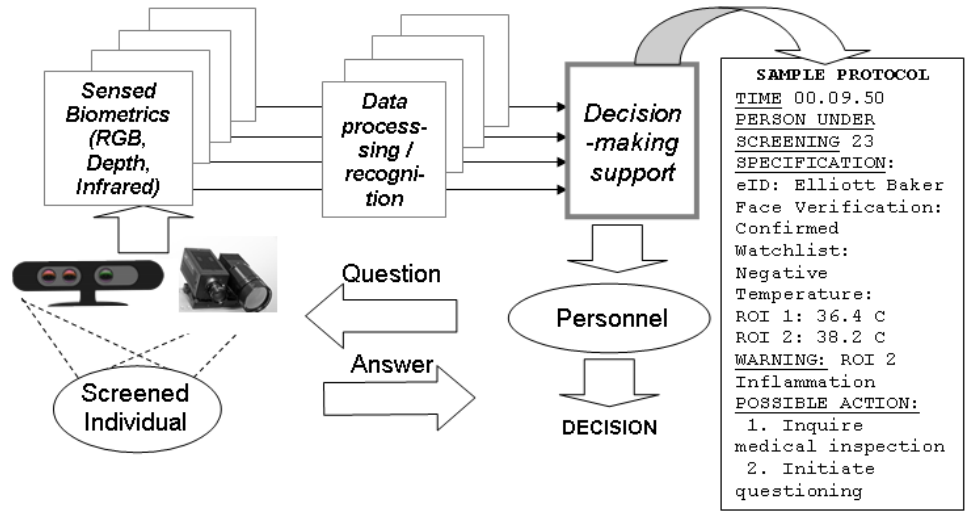}
		 \caption{The access control system that uses facial biometrics across various ranges and supports the personnel in the authorization of
individuals through screening an interview.}\label{fig:screening-discipline}

\end{figure*}

A particular structure of a screening and interview support system  is shown in Fig. \ref{fig:screening-discipline}. The system consists of the \emph{cameras} in RGB-Depth (RGB-D) and infrared bands, the \emph{processors} of preliminary information and online data (such as watchlists), and decision-making support that include converting the results into a semantic form such as a protocol made available to the personnel. This allows for  \emph{dialogue support} in order to assist personnel in conversation with a customer on the topic of   an authorization. This semantic data is based on the preliminary  information gathered from surveillance of the screened person in the visible and infrared bands, and information extracted from observation, conversation, and  additional sources.  The process of generating questions is initiated by  information sensed by biometric devices.  The questionnaire strategy can  alleviate some  errors, as well as  unreliability of biometric data. 

For example, the description from the authorized documents may not match the appearance of the individual, and facial recognition may provide the required support. The interview-support system may prompt a dialog in order to clarify the situation.

The other sensory data, such as infrared, may assist in fever screening. In \cite{[kn:Ng-infrared-MassBlindFeverScreen]}, infrared mass blind screening of potential fever subjects such as SARS or bird flu patients  was studied. A handheld radiometric infrared ThermaCAM S60 FLIR system was used. In experiments, the focal length from individual to scanner was 2 meters with a duration of scanning of 3 seconds. 

In an interview-supporting system, if high  temperature in an individual is detected and reported to the personnel, one of the automatically generated questions suggested for the officer to ask  this individual might be as follows: \vspace{3mm}

\begin{center}
\mbox{\begin{minipage}{8.5cm} {
\begin{ttfamily}
\begin{footnotesize}
\underline{Sample Protocol}\\
POSSIBLE ACTION: 
Inquire: Have you been experiencing a high fever? 
\end{footnotesize}
\end{ttfamily}}
\end{minipage}}
\end{center}
\vspace{3mm}

Another focus of interest in such a system includes artificial materials (fake moustache, makeup, wig, etc.) and surgical alterations (plastic  surgery technology). The infrared spectrum data provides useful information for detection of disguised features \cite{[kn:Bhanu-Pavlidis2005]}.

Unreliable data on artificial accessories in an infrared facial image are transferred into a semantic form of the protocol as follows: 
\vspace{3mm}
\begin{center}
\mbox{\begin{minipage}{8.5cm} {
\begin{ttfamily}
\begin{footnotesize}
\underline{Interaction Protocol}\\
WARNING: Possible intention to change  appearance; features of
artificial moustache are detected.
\end{footnotesize}
\end{ttfamily}}
\end{minipage}}
\end{center}
\vspace{3mm}


The personnel can, in addition to the automatic aid, analyze the acquired raw images in the visible and infrared spectra. The individual's facial (RGB) data are verified against the data in eID (such as e-passport) and also compared against
global databases (such as watchlists). Note that data on an individual may not always be available in the database - this is the worst case scenario, and  intelligent support is vital in this case.

\section{Facial biometrics in access control: case study}

Facial biometrics used in border control now, as well as ones planned for use in the next decade, are primarily  based on 2D face recognition technologies as per the recommendations of the ICAO \cite{[kn:ICAO-e-passport]}. This is because the face templates, such as the ones used in e-passports, are based on 2D photos. Australia, New Zealand and most European countries are now using face-biometric enabled gates (e-Gates) for border control, which includes the e-passport/e-ID reader  and the RGB camera. e-Gates verify the passengers biometric data (facial image) against the travel document, such as an e-passport, e-ID, or a pre-existing database containing biometric data \cite{[ABC-IATA]}.

The disadvantages of today's 2D facial recognition techniques are well known. The best results of the 2D facial recognition algorithms tested on the 2010 benchmark data set are reported in \cite{NIST_Evaluation_2D_Still-Image_Face} as follows: a FNMR of 0.3\% at a FMR of 0.1\%. However, these results do not conform with the worldwide border control statistics since the actual environment has other factors, such as airport logistics and human factors (untrained users), that cause significant performance degradation. The table below shows the results of the operation of the EasyPASS system (Germany) as reported in \cite{[kn:EasyPASS-II],[kn:EasyPASS-I]}.

\begin{table}[!hbt]
\caption{Sample of statistical data from  the border  control
system EasyPass based on facial recognition (Oct.2009 -
Sept.2010) \cite{[kn:EasyPASS-II],[kn:EasyPASS-I]}}
\label{tab:EasyPASS111}
\begin{small}
\begin{center}
\begin{tabular}{l|l}
\hline \multicolumn{1}{c|}{\textbf{Statistical parameter}} &
 \multicolumn{1}{c}{\textbf{Estimated value}}
  \\\hline
\begin{parbox}[h]{0.5\linewidth}{
\vspace{1mm} Total number of users
 \vspace{1mm}}
\end{parbox}&
\begin{parbox}[h]{0.4\linewidth}{
\vspace{1mm} 50.000 \vspace{1mm}}\end{parbox}
\\   \hline
\begin{parbox}[h]{0.5\linewidth}{
\vspace{1mm} Success rate
 \vspace{1mm}}
\end{parbox}&
\begin{parbox}[h]{0.4\linewidth}{
\vspace{1mm} 86\%$^{\mathbf{*}}$ \vspace{1mm}}\end{parbox}
\\   \hline
\begin{parbox}[h]{0.5\linewidth}{
\vspace{1mm}  Rejection rate
 \vspace{1mm}}
\end{parbox}&
\begin{parbox}[h]{0.4\linewidth}{
\vspace{1mm} 14\% \vspace{1mm}}\end{parbox}
\\   \hline
\begin{parbox}[h]{0.5\linewidth}{
\vspace{1mm}  Rejection due to face
  verification failure
 \vspace{1mm}}
\end{parbox}&
\begin{parbox}[h]{0.4\linewidth}{
\vspace{1mm} 5.5\% at 0.1\% FAR \vspace{1mm}}\end{parbox}
\\   \hline
\begin{parbox}[h]{0.5\linewidth}{
\vspace{1mm}  Period to pass the e-gate$^{\mathbf{**}}$
 \vspace{1mm}}
\end{parbox}&
\begin{parbox}[h]{0.4\linewidth}{
\vspace{1mm}   18 seconds \vspace{1mm}}\end{parbox}
\\   \hline
\begin{parbox}[h]{0.5\linewidth}{
\vspace{1mm}  Recognition performance
 \vspace{1mm}}
\end{parbox}&
\begin{parbox}[h]{0.4\linewidth}{
\vspace{1mm}   5\% FRR at 0.1\% FAR \vspace{1mm}}\end{parbox}
\\   \hline
\end{tabular}\\ 
\end{center}
\end{small}
\end{table}

Based on these statistics, the number of people who were directed to the manual inspection lane at EasyPASS can be estimated as
$ {50,000}-
 {43,000} =
   {7,000}$.
This means that every 7th traveler $(50,000/7,000\approx 7)$ was directed to the manual inspection lane.
Statistics provided in \cite{[kn:Cantarero-2013]} for 2013, show that the best facial FRR has been  achieved by the border control systems located in Portugal airports (FRR=1.36\%), and the worst result (FRR=18.18\%) was demonstrated by the systems in Denmark. In total, 1 out of 10 travelers was directed to manual control because of facial verification failure (FRR=9.30\%, given 18,884 travelers who used the system).

Meanwhile, the advances in  3D face recognition  provide an improved robustness while processing variations in poses and problematic lighting conditions. However, this technology  relied on expensive 3D scanners until recently. Inexpensive near-infrared range sensors such as one developed by PrimeSense and the now familiar Microsoft Kinect sensor, have become available since 2010. Safran SA has developed a Morpho 3D Face Reader that uses a similar principle. 

In this paper, we argue that advances in range and sensory technology must be utilized to support the existing multi-spectral facial biometric data accumulation to support the authentication process. In particular, we consider how depth data acquired by the range sensors, such as the Kinect, can facilitate facial recognition and facial  temperature reading, thus assisting the decision-making in authentication  and risk management (such as fever screening).

\section{RGB and Depth Facial Biometrics}
\label{s.propose}

Below we evaluate the possible inexpensive enhancement of face verification using RGB-D sensors (RGB and Depth data) instead of traditional RGB (regular) video cameras. This approach does not require storing 3D templates (as in the case of purely 3D face recognition), neither does it require usage of time-consuming 3D data processing, or time-consuming geometrical transformation when using a multi-camera (stereo-camera) setup.  

We propose to use a fast head pose estimation algorithm that performs real time image processing using a RGB-D camera. In particular, any of the current Kinect, Carmine or Asus sensors supply both depth data and RGB videos, providing a resolution of $640 \times 480$. Instead of using Depth data for retrieval of 3D data, it can be used to perform pre-processing and detection. After that, a 2D recognition can proceed. 

Note that, in addition, the Depth sensor can be instrumental in verifying the liveness of the object in order to prevent tampering by presenting the sensor with a photo or a smart phone image of a person.

\subsection{Head Pose Estimation}
The head pose estimation algorithm is based on the work of Fanelli et. al \cite{Fanelli} where depth data is acquired in order to estimate the head orientation using a random regression forest implementation. This approach is applied to find the most optimal position of one's face.  

\begin{figure}[!ht]
      \includegraphics[width=7cm]{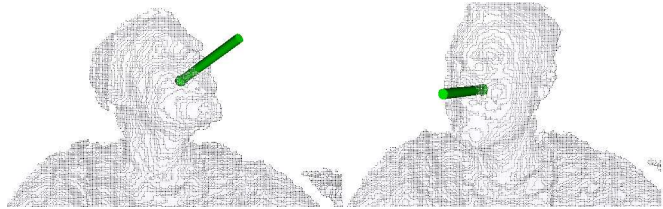} 
	\caption{Head-Pose Orientation Vector}
\label{f.g1}
\end{figure}

Figure \ref{f.g1} displays the 3D model of a human subject using the depth information from the Kinect camera. The green cylinder represents the orientation vector of the person and  is drawn by connecting the center of the head to the nose.  Both the center of the head and nose are calculated by the fitting of a 3D mask to the depth model; the mask is created based on the random regression forest algorithm \cite{Fanelli}. After computing the head-pose vector, the latter is compared with the $z$-axis using the dot product. This comparison yields the offset angle between the vectors. Using the offset angle, it is possible to filter frontal view images within a video. Obtaining the frontal view images is crucial for the facial recognition algorithm  accuracy. Also, by selecting the first occurring frontal view from video, it is possible to perform faster facial recognition compared with methods that run recognition continuously throughout a video sequence to ensure good recognition rates.

\subsection{Face Detection via Pose Estimation}
The most popular approach to detection of the face in an image is the Haar-like feature detection, implemented in OpenCV \cite{[opencv]}. It involves finding a region of interest (ROI), and then the use of a sliding window algorithm that slowly grows in size up to a specified threshold. For each ROI, the feature detection is run through a cascaded classifier. 

\begin{figure*}[!hbt]
\hspace{12.5mm}
      \includegraphics[width=13.8cm]{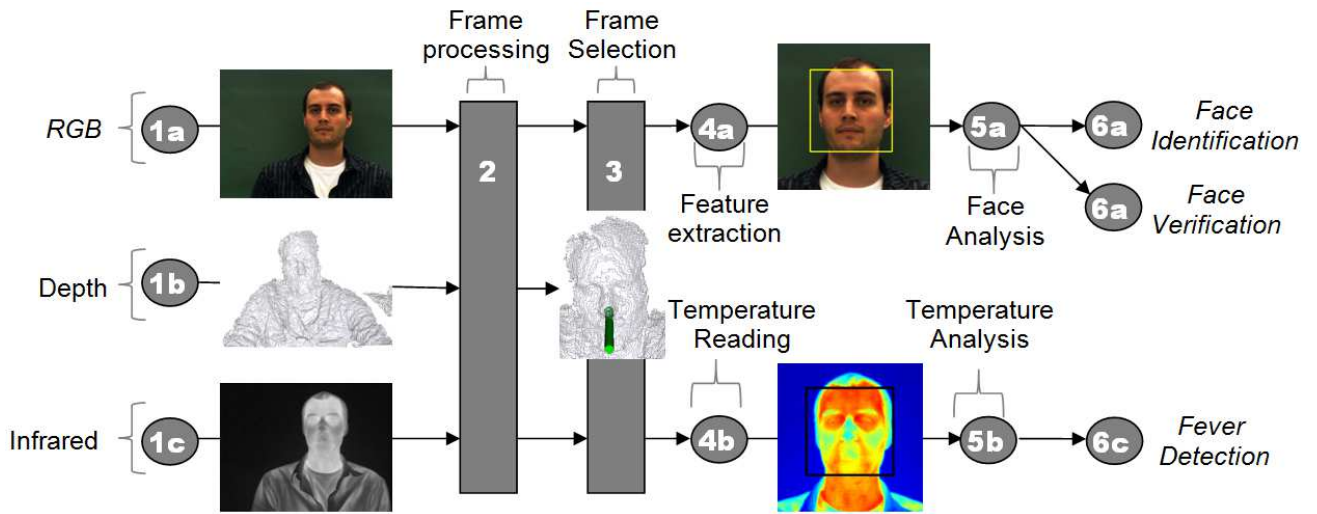} 
	\caption{Mutli-spectral analysis of face biometrics.}
\label{f.g2}
\end{figure*}

Another form of face detection is to use the head-pose estimation algorithm. The pose estimation algorithm detects the head orientation vector, as well as the location of the head.  Based on this data, the face can be cropped and used for facial recognition. It  also assists the task of infrared image processing of the face for the purpose of region-of-interest detection and further facial temperature estimation. In our recognition system we replace the detection algorithm with head-pose estimation. Figure \ref{f.g2} illustrates the entire algorithm.

\subsection{Face Recognition and Verification}
 In this paper, the facial recognition is implemented using the FaceRecognizer class in OpenCV \cite{[opencv]}.  Three main recognition algorithms have been evaluated:
\begin{itemize}
	\item EigenFace
	\item FisherFace
	\item Local Binary Patterns Histograms (LBPH)
\end{itemize}

\section{Infrared Facial Biometrics}

Biometric technology, that uses infrared thermography, descends  from medical applications, namely, diagnostic methods that provide information about normal and abnormal functioning of the sensory and sympathetic nervous system, vascular dysfunction, myofascial trauma, and local inflammatory processes.

The infrared image analysis   includes recording an infrared video image, infrared image processing, and evaluation; in particular, of temperature and blood flow rate. The fluctuation of temperature in various facial regions is primarily due to  changing blood flow rate. In \cite{[kn:Fujimasa00]},  heat-conduction formulas at the skin's surface are introduced. The thermodynamic relation between the blood flow rate $V_S$  at the skin level, blood temperature at the body core $T_{blood}$, and  skin temperature $T_{skin}$  are used to convert infrared intensity to temperature ${dV_s}/{dt}=f(T_{blood},T_{skin}).$ By solving this for every point in the image,  raw thermal data is transformed into blood flow rate data.

In medical applications, infrared-based diagnostic systems provide an accurate quantitative analysis of temperature distribution on a target surface: absolute and  mean temperature of any region; in particular, of the face, and differences between the right and the left sides of the face.

A breath-monitoring  function has been well studied in polygraph usage \cite{[kn:Polygraph-Lie-Detection]}. The distance infrared  breathing measure function is based on the fact that exhaled air has a higher temperature than the typical
background of indoor environments \cite{[kn:Murthy-Pavlidis]}.

%

In the context of access control systems that perform multi-spectral analysis of human biometrics, temperature is an important source of information. For example, the systems with situational awareness and monitoring of travelers in border control shall have non-invasive remote temperature measurement using IR cameras for fever detection. 

In our research, head detection and head pose estimation based on RGB-D data, specifically the frontal view detection, assists another task,  - estimation of facial temperature based on the infrared  (IR) images as depicted in Figure \ref{f.g2}. Using an IR image of a frontal-view face,   fever and/or high temperature can be detected and the system shall alert the  personnel on such findings.  The RGB-D image processing is used  to detect not only the RGB  images of the frontal view but also to extract the synchronized frames from IR video. Next,  the regions-of-interest are detected, and the IR image intensity is further processed to be converted to temperature measurements. In addition, blood flow rate can be estimated when the skin/body temperature is calculated from the sequence of IR frames \cite{[kn:Fujimasa00]}.

\section{Experimental Results}
\label{s.exp}
The experiments are conducted on an Intel\textsuperscript{\textregistered} Core\texttrademark{} i7 Q740 (1.73GHz) computer with 8GB of RAM. We collected a Local database containing  RGB and depth videos with 17 different individuals as well as a synchronized IR video of the same subjects.

For the pose estimation and face recognition experiments, we used ORL \cite{Samaria}, ICT-3D HeadPose \cite{Fanelli}, and Biwi Kinect Head Pose databases \cite{Fanelli2}.  

The {Biwi Kinect Head Pose Database} has a total of 20 subjects and approximately 500 RGB images at a resolution of 640x480 pixels with depth information for each subject. Each subject's pose range varies about $\pm75^\circ$ yaw and $\pm60^\circ$ pitch.  For each image a text file providing the ground truth head rotation and location was given for comparison.

The {ICT-3D HeadPose Database (ICT-3DHP)} includes datasets of video that include both RGB and Depth collected using  Microsoft Kinect \cite{Baltrusaitis}.  The database contains 10 RGB videos with subjects rotating their heads at varying yaw, pitch, and roll angles.  For every RGB frame in the video, there is a depth image.  In addition, each dataset is labeled with ground truth head poses obtained by using the Polhemus FASTRACK flock-of-birds tracker.  

The {ORL Database of Faces} contains RGB images of different subjects taken at different time intervals by \cite{Samaria}. There are a total of 40 subjects with 10 256 gray level 92x112 pixel images for each. 

The local lab database contains multiple videos obtained using a Kinect\textsuperscript{\textregistered} sensor on 17 different subjects.  For the first 16 subjects, they are tasked with 2 different roles:  rotating head from left to right twice, and change facial expression (neutral, anger, sad, happy, surprise, disgust and fear).  The last subject (17th) performs one extended head rotation consisting of left to right, clockwise, counter-clockwise, and up to down.  Each task is divided into 3 sets of video: 16 rotation videos (VS1), 16 facial expression videos (VS2), and one extended head rotation (VS3).  Each frame (both depth and RGB) is 640x480 pixel resolution; every 15th frame is used instead of every frame to provide a compromise between the overall consistent video flow and distinctive sequential frame variance.

The purpose of the first experiment was to evaluate the accuracy of the orientation of the head using the extended head rotation video. Each frame that is determined to have an optimal pose is saved. The extended rotation video in the local database (VS3) was used for this experiment. Also, the Biwi database was examined to demonstrate the accuracy between the ground truth and experimental frontal selection.

\begin{figure}[!ht]
 \includegraphics[width=8cm]{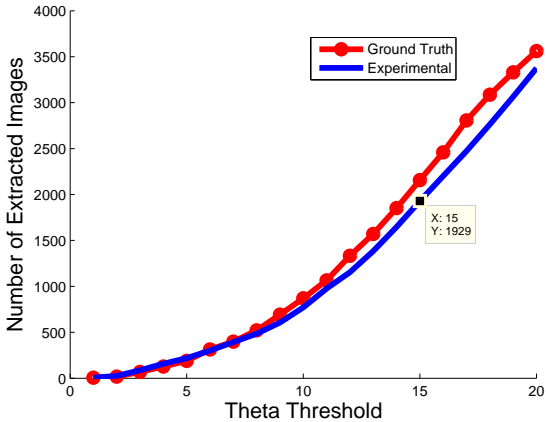} 
\caption{Varying Theta Threshold on Biwi Database.}
	\label{f.threshold}
\end{figure}

Figure \ref{f.threshold} displays the curve representing the number of frames extracted based on varying the angular (Theta) threshold.  The selected point on the experimental curve shows 1929 extracted faces at $15^\circ$ threshold.

\begin{figure}[!ht]
      \includegraphics[width=8cm]{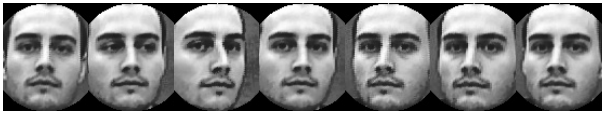}  
	\caption{ 7 accepted frames from video using head estimation algorithm with $15^\circ$ threshold.}
\label{f.poseFrames}
\end{figure}

This estimation algorithm was trained using the Biwi database  \cite{Fanelli} and the angular threshold of $15^\circ$ was adopted for the extended video (VS3) experiment which is displayed in Figure \ref{f.poseFrames} for frames accepted within the threshold .

The second experiment was to investigate face detection; it was performed using 16 local videos (VS1) and the following face detection approaches:
\begin{itemize}
	\item Haar-like (Viola-Jones algorithm) implemented in OpenCV \cite{[opencv]};
	\item Haar-like depth head pose (HL*DHP) developed in this study.
\end{itemize}

\begin{table}[!htbp]\footnotesize 
		\caption{Face Detection on Different Databases} 
	\centering
		\begin{tabular}{|c|c|c|c||c|}
		\hline
		 & {\bf Rate} & {\bf Face}& {\bf Optimal}& {\bf Total}\\
		 &  & {\bf Detected}& {\bf Frames}& {\bf Frames}\\
		\hline
		\hline
		{\bf Local}&&&&\\
		\hline
		Haar-Like					& 48.20\% 		&		191	&	-		& 429\\ 
		HL*DHP						& 83.33\%			&		65	&	89	&	330\\
		\hline 
		\hline 
		{\bf Biwi} &&&&\\
		\hline
		Haar-Like					& 46.73\% 		&		7326	&	-		& 15677\\ 
		HL*DHP						& 97.17\%			&		1889	&	1944	&	15677\\		
		\hline
		\hline 
		{\bf ICT3DHP} &&&&\\
		\hline
		Haar-Like					& 58.77\% 		&		8352	&	-		& 14212\\ 
		HL*DHP						& 80.20\%			&		5325	&	6640	&	14212\\		
		\hline
		\end{tabular}
\label{t.faceDetection}
\end{table}

Table \ref{t.faceDetection} shows the face detection rates on a set of 16 videos (VS1). The face detection rate is measured as a ratio between the number of detected faces and the number of frames (total or optimal, which represents the number of frontal views). As there is no negative inputs (every frame contains a face), false positive rate is not considered. The Haar-like method runs a face detection program without any depth assistance, thus no optimal frames are extracted.  HL*DHP method incorporates both the HL and DHP method; the DHP assists the HL method in selecting the optimal frames before applying face detection. The optimal frames are selected based on the angular threshold for extracting the best frontal views. As follows from Table \ref{t.faceDetection}, applying the DHP greatly increases the face detection rates since the head estimation algorithm can detect the frontal views of a head. The overall 14\% difference between local HL*DHP and Biwi HL*DHP is in favor of the latter due to the much more robust selection of parameters and background-subtracted dataset in the Biwi database.  The face detection rate of 58.77\% for ICT-3DHP shows that this database's video sequences are composed of more frontal views as opposed to Biwi sequences, showing the detection rate of 46.73\%.  The gain from  58.77\% to 80.20\% for ICT-3DHP due to incorporation of head pose estimation is lower than for the other databases, which can be explained by the fact that the ICT-3DHP contains a larger proportion of frontal views.

The third experiment investigated face matching in order to  evaluate the efficiency of the proposed approach. For the ORL database, 5 images from each of 40 subjects were used to train the algorithm and 5 images per subject were used to test it.  For the local database, local expression videos (VS2) were used for the training set while local rotation videos (VS1) were used for testing.  For the Biwi database, 1 image per subject was extracted for training and the remaining images were used for testing.  For the ICT-3DHP database,  1 image per subject was used for training and the remaining images were used for testing.

The comparison process involves face recognition and verification. 
Table \ref{t.2} shows the results of the three recognition methods implemented in OpenCV using ORL, Local, Biwi, and ICT-3DHP databases.  

\begin{table}[!htbp]\footnotesize 
		\caption{Recognition Accuracy with Varying OpenCV methods and Databases} 
	\centering
		\begin{tabular}{|c|c|c|c|}
		\hline
		 & \multicolumn{3}{c|}{\bf Methods} \\
		 \hline
		 & { EigenFace} & {FisherFace}& { LBPH}\\
		\hline
		\hline
		{\bf ORL}&&&\\
		\hline
		Traditional				& 92.0\% 		&		91.5\%		&	88.5\%			\\ 
		Proposed						& -		&		-	&	-	\\
		\hline
		\hline
		{\bf Local}&&&\\
		\hline
		Traditional				& 76.14\% 		&		78.48\%		&	84.56\%			\\ 
		Proposed						& 85.33\%			&		85.52\%		&	91.52\%		\\		\hline 
		\hline 
		{\bf Biwi} &&&\\
		\hline
		Traditional				& 63.77\% 		&		63.77\%	&	63.72\%		\\ 
		Proposed						& 88.30\%		&		88.30\%	&	88.25\%	\\		
		\hline
		\hline 
		{\bf ICT-3DHP} &&&\\
		\hline
		Traditional					& 60.48\% 		&		60.48\%	&	60.44\%		\\ 
		Proposed						& 70.14\%			&	70.14\%	&	68.96\%	\\		
		\hline
		\end{tabular}
\label{t.2}
\end{table}

Table \ref{t.2} shows the results of comparison of the traditional approach to  face recognition and  detection for the RGB videos with no head-pose/depth assistance, and the proposed approach that uses head-pose/depth assistance prior to detection/recognition on the RGB videos. For the traditional approach, the accuracy of the recognition methods are measured as a ratio between the number of correct recognitions and the total number of detected faces.  For the proposed approach, accuracy is measured as a ratio between the number of correct recognitions and the total number of faces detected after selecting the best frontal view with a $15^\circ$ threshold.  Since there is only one output per image, the accuracy measurement accounts for false positives (the face of person A is recognized as the face of person B).

It follows from Table \ref{t.2} that the recognition rate increases about 6-9\% for the local database by just incorporating a head pose estimation algorithm to find the best frontal view before applying face detection and recognition.  For the ICT-3DHP database, the recognition accuracy is about 10-15\% lower than for the other databases; however, the overall gain in performance by incorporating head pose estimation is 10\%.  In addition, the recognition rates show about 25\% gain for the Biwi database.  For the Biwi and ICT-3DHP database, the same approach demonstrates similar accuracy for different recognition methods because only one image was used for training.  Increasing the number of images to train each recognition method will increase the accuracy for both traditional and the proposed approach while creating more disparity between each recognition method. However, only one image, the frontal view, was specifically chosen for training to demonstrate the effectiveness of both head-pose estimation and frontal view recognition.  

In the case of Biwi, the traditional approach extracts  faces, both frontal and rotated ones, which are then used for facial recognition.  The recognition accuracy in this approach is 63.7\% because the rotated views are not used for training.  On the contrary, the proposed approach demonstrates reasonable accuracy, 88.3\%, as this approach extracts near-frontal views which contain enough similarity to the trained image to be correctly recognized.

In the case of ICT-3DHP, the experimental data shows an overall lower recognition accuracy than for a similar approach reported in Biwi experiment.  The drop in accuracy is likely due the slight difference in the quality of the recorded video sequences; the parameters such as lighting and distance from camera are different.

Face verification  either assumes that the enrolled  images are the e-passport photo or an image in the  existing database.  By comparing the enrolled database with live images a verification is performed to determine the validity.  For this experiment, the first three frontal view images per subject are used as enrollment, or training set, and the other images are used as test.

\begin{figure}[!ht]
  \includegraphics[width=9cm]{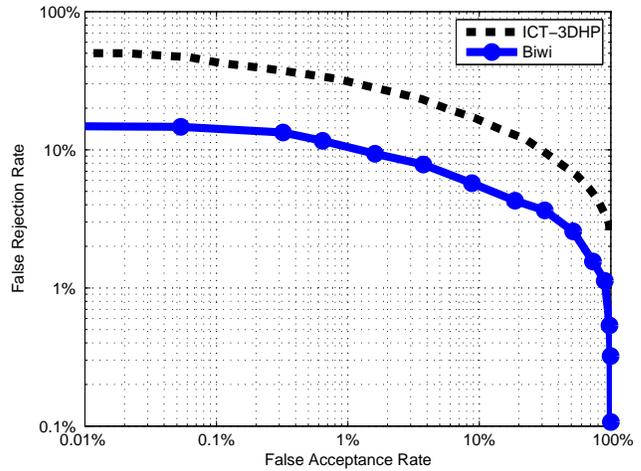} 
	\caption{FRR vs FAR using Verilook \cite{[verilook]}.}
	\label{f.roc}
\end{figure}

Table \ref{t.faceVer} and Figure \ref{f.roc} show the results of the face verification using the proposed method on the Biwi and ICT-3DHP databases.  The proposed approach uses the DHP to filter out the non-frontal view images.  Results in Table \ref{t.faceVer} are relatively similar to those with the E-pass gate performance statistics. Therefore, RGB-D cameras may be a replacement to standard e-Gate cameras with the benefit of faster processing.
\begin{table}[!htbp]\footnotesize 
	
	\caption{Face Verification using Verilook SDK \cite{[verilook]} on Different Databases} 
	\centering
		\begin{tabular}{|c|c|c|}
		\hline
		& {\bf EER} &{\bf FRR @0.1\% FAR}\\
		\hline
		{\bf Biwi} 														& ~6.5\%		& ~10.1\%\\		
		\hline
		{\bf ICT-3DHP} 												& ~14.2\%		& ~30.2\%\\		
		\hline
		\end{tabular}
\label{t.faceVer}
\end{table}

Table \ref{t.Time} compares the time required to perform recognition, DHP, detection, and processing for both approaches using the local rotation video (VS1).  For all 16 videos, every procedure is timed then averaged based on the number of frames per video.  For each procedure, the following timings are measured on a per-frame basis:

\begin{itemize}
	\item Recognition: Time spent on predicting target;
	\item DHP: Time spent on calculating head-pose vector;
	\item Detection: Time spent on detecting face;
	\item Processing: Total Time spent.
\end{itemize}

\begin{table}[!htbp]\footnotesize 
	
	\caption{Operation Time on Traditional VS. Proposed Method} 
	\centering
		\begin{tabular}{|c|c|c|c||c|}
				\hline
		{\bf Method}   & {\bf Recognition} &  {\bf DHP }&  {\bf Detection }&  {\bf Processing }\\
		\hline
		\hline
		{\bf Per Frame}    &{\bf (ms)}&  {\bf (ms)}&  {\bf (ms)}&  {\bf  (ms)}\\
		\hline
		Traditional	&  37.09	& -			& 226.52	& 414.72 \\
		Proposed	& 32.84	& 113.58 & 203.89 & 280.30 \\
		\hline
		\hline
		{\bf Per Video}    &{\bf (s)}&  {\bf (s)}&  {\bf (s)}&  {\bf  (s)}\\
		\hline
		Traditional	&  0.59	& -			& 3.62	& 6.64 \\
		Proposed	& 0.38	& 1.32 & 2.40 & 3.28 \\
		\hline
		\hline
		\hline
		{\bf Total}    &{\bf (s)}&  {\bf (s)}&  {\bf (s)}&  {\bf  (s)}\\
		\hline
		Traditional	&  6.6	& -			& 92.05	& 168.08 \\
		Proposed	& 2.30	& 34.19 & 16.88 & 88.64 \\
		\hline
		\end{tabular}
\label{t.Time}
\end{table}

Table \ref{t.Time} shows the operation per-frame time based on both the traditional and optimized approach.  The processing per-frame time is about 134 ms faster for the optimized approached when compared to the traditional approach.  This gain is explained by incorporating the DHP algorithm which reduces the amount of time the face detection algorithm needs to be run.  Since the DHP is a faster algorithm, compared to face detection, it can be used as a filter in order  to reduce the total number of times the face detection algorithm is required for an entire video by only using the frontal view frames.  Based on Table \ref{t.Time}, there is a 32.44\% gain in speed.  However, in order to enable the DHP algorithm,  1.13 seconds are required for loading and training. 


In addition, we conducted experiments on  temperature reading from the frames of the IR video, that were synchronized with the RGB frames extracted using the depth-based head position estimation described above. We used the infrared camera Miricle 307k from Thermoteknix. Camera calibration was performed using a thermometer and  defining the pixel intensity and the corresponding temperature. 

In order to select a region of interest within an IR  image either  manual or automatic selection can be used.  For automatic selection, a face detection algorithm is applied on the RGB frames; next, a corresponding IR frame is also selected. Superimposing  both frames enables detecting the head on the IR image and, using the head proportions, selecting the regions-of-interest such as the forehead.



The approximation is performed by measuring different temperatures at different pixel intensities.  Using the vectors of temperature with respect to intensity, a graph representing the relationship is drawn (Figure \ref{f.temp}) where the linear expression is used to estimate temperature shown below:

\begin{equation}
y = 0.2087x + 22.28
\end{equation}

where $y$ is the estimated temperature, $x$ is the averaged gray level intensity of bounding box, 0.2087 is the slope found in the graph, and 22.28 is the y-intercept and also the calculated room temperature.  More advanced and accurate approaches such as \cite{[vidas]} can be used.  Following \cite{[kn:Fujimasa00]}, the temperature can be used to calculate the blood flow rate.  

\begin{figure}[!ht]
\hspace{3mm}
\begin{tabular}{ccc}
			\includegraphics[width=2.1cm]{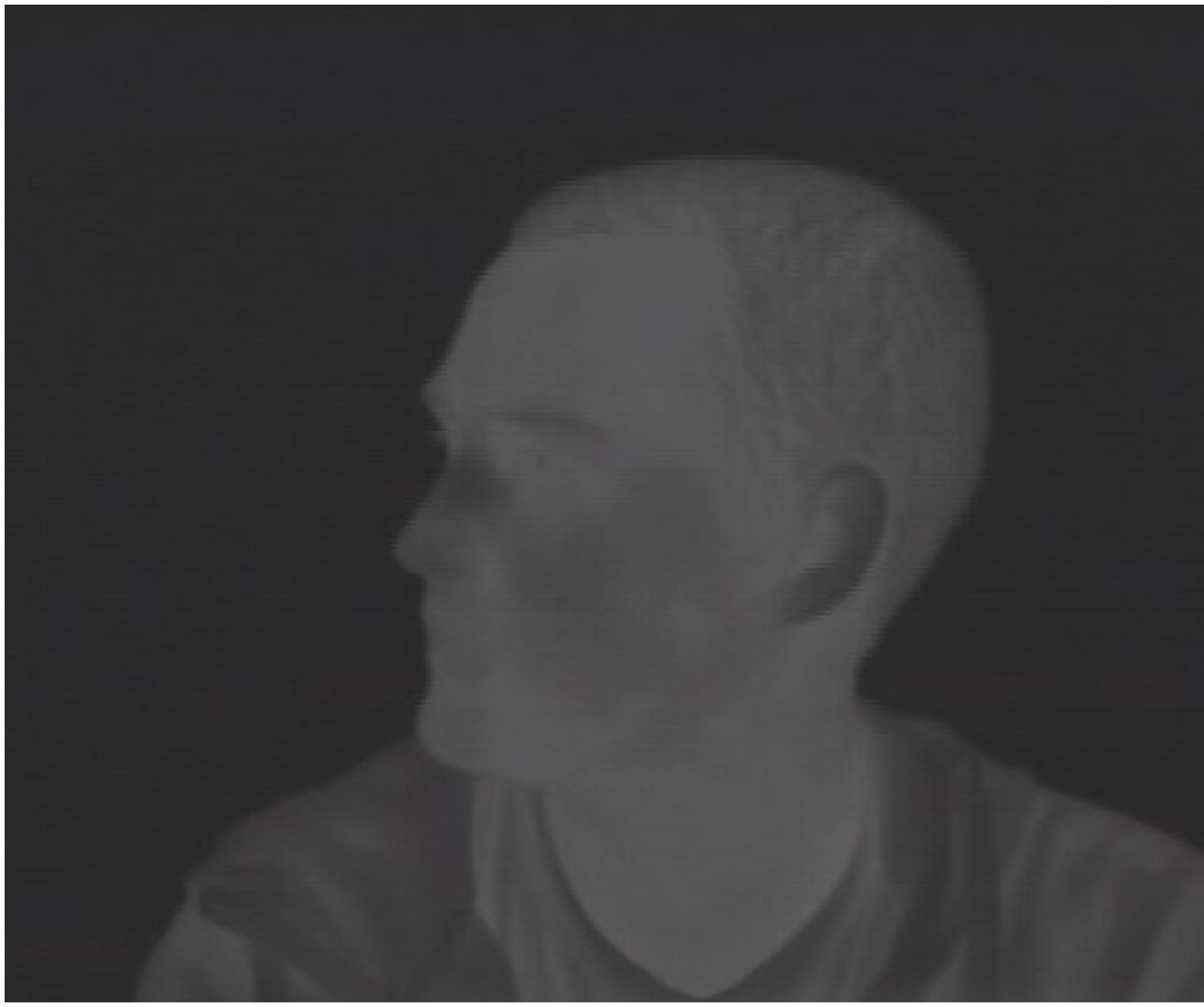}\hspace{3mm} &
			\includegraphics[width=2cm]{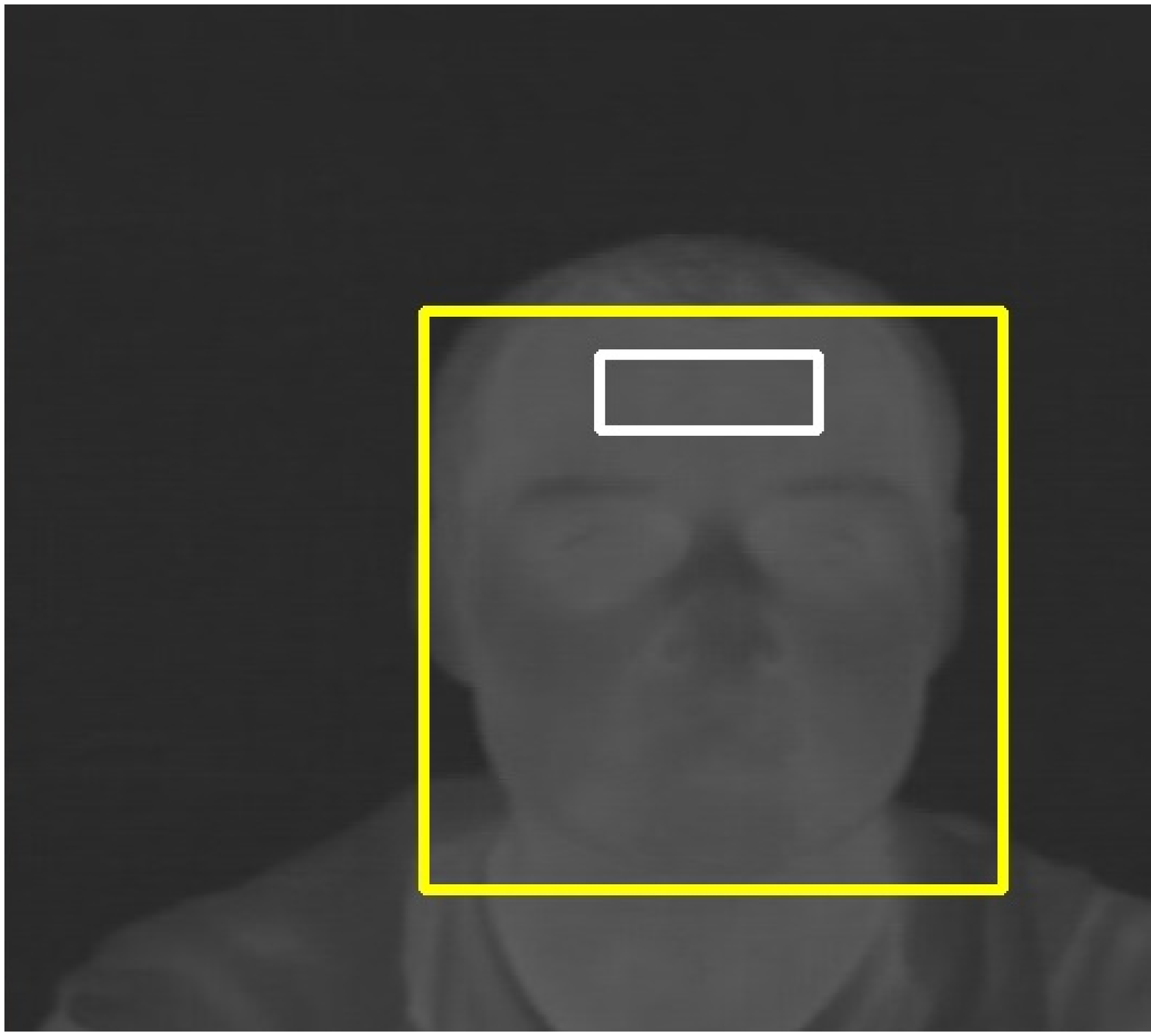}\hspace{3mm} &
			\includegraphics[width=2.1cm]{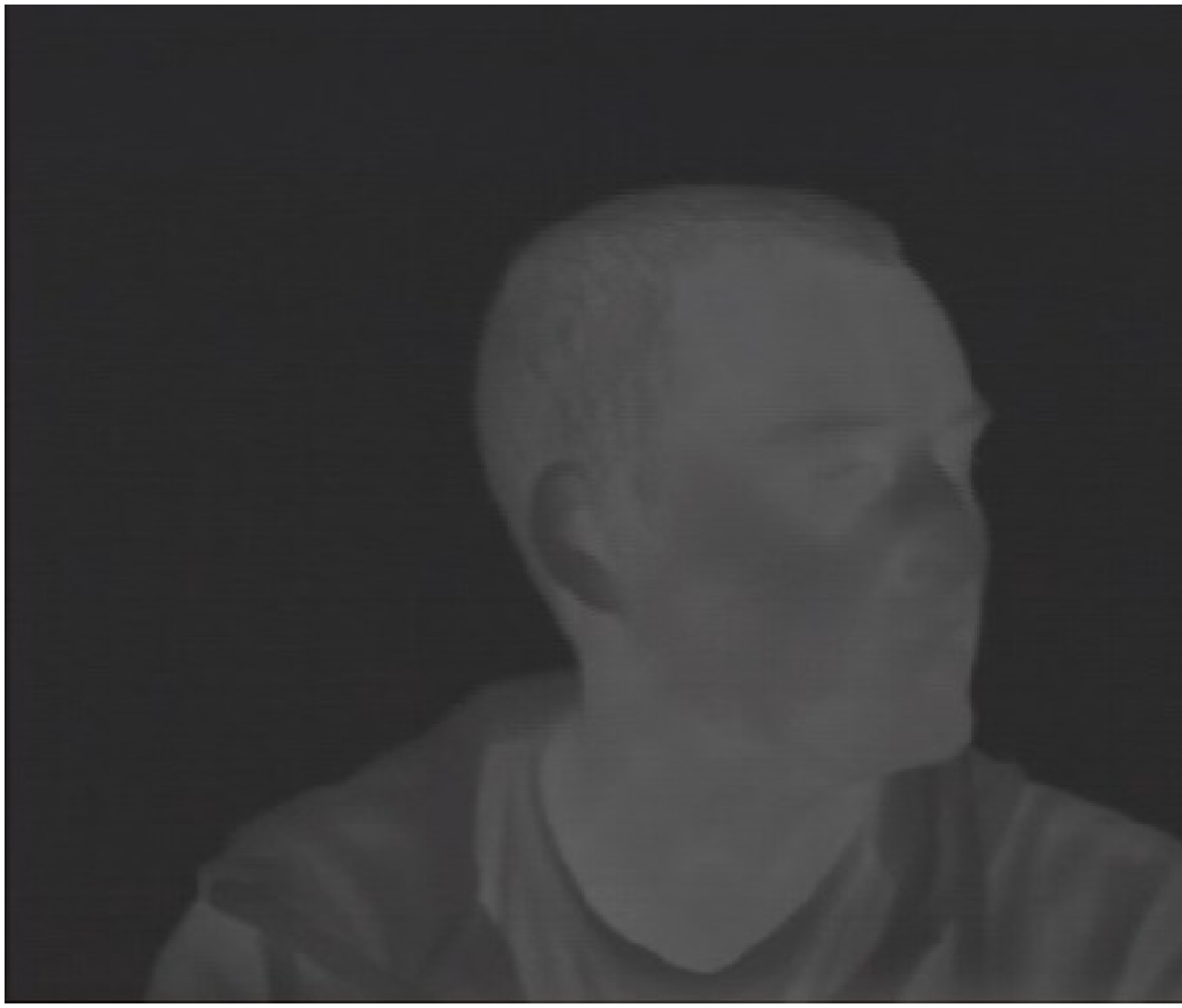} \\
			\includegraphics[width=2.1cm]{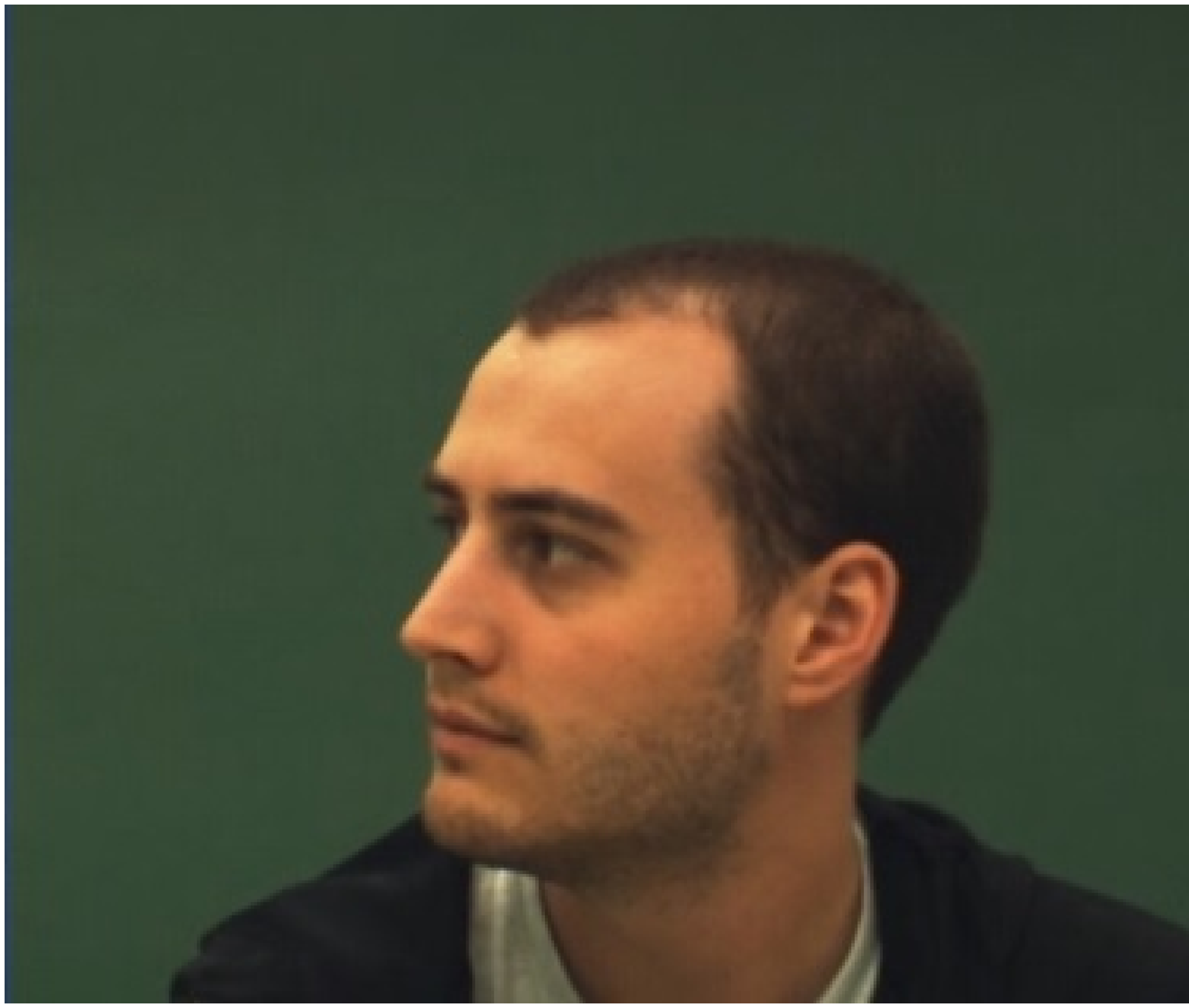}\hspace{3mm} &
			\includegraphics[width=2cm]{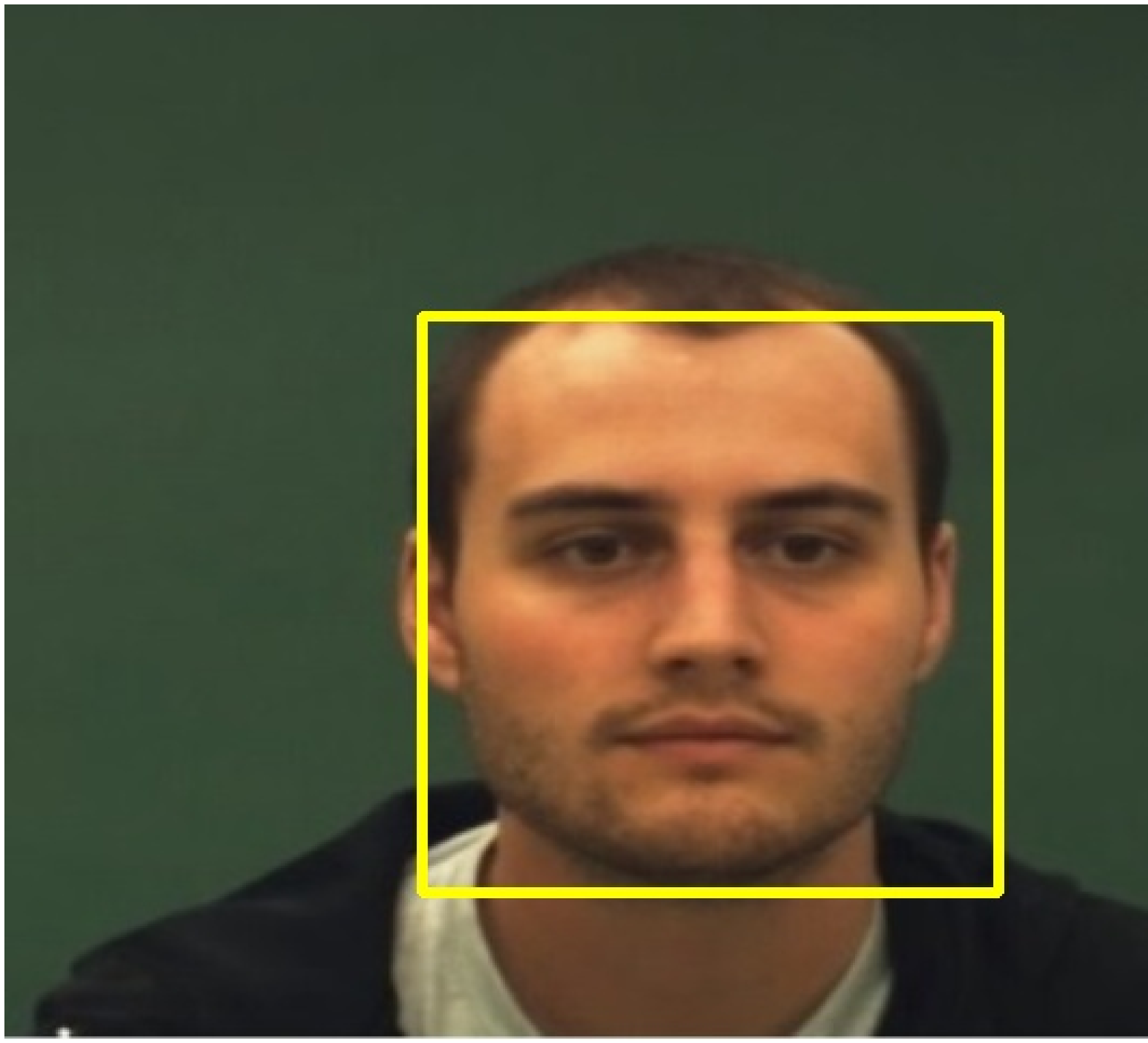} \hspace{3mm} &
			\includegraphics[width=2.1cm]{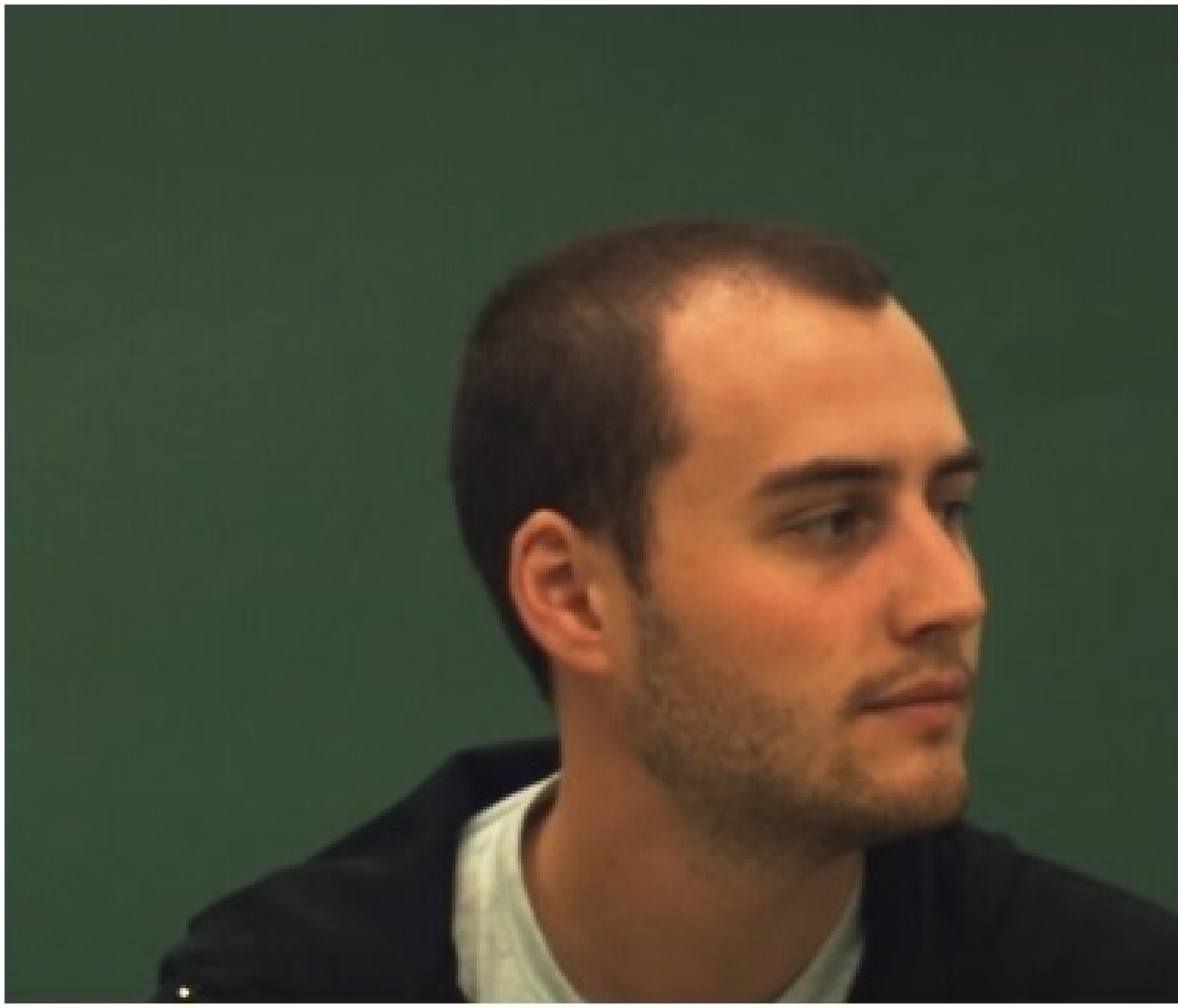} \\
      (a) & (b) & (c)
  \end{tabular}

	\caption{RGB and Gray-scale IR image: (a) Left Rotated Face that failed pose estimation; (b) Frontal Face that passed pose estimation and thus returned with temperature information: forehead temperature of 33.727$^\circ$C with a estimated blood flow rate of 39.6536 $ml/100g tissue\cdot min$ and an overall head temperature of 31.5156$^\circ$C with a estimated blood flow rate of 19.2156 $ml/100g tissue\cdot min$; (c) Right Rotated Face that failed pose estimation.}
	\label{f.t1}
\end{figure}

\begin{figure}[!ht]
	
	\begin{center}
	\begin{tabular}{cc}
	\includegraphics[width=4.15cm]{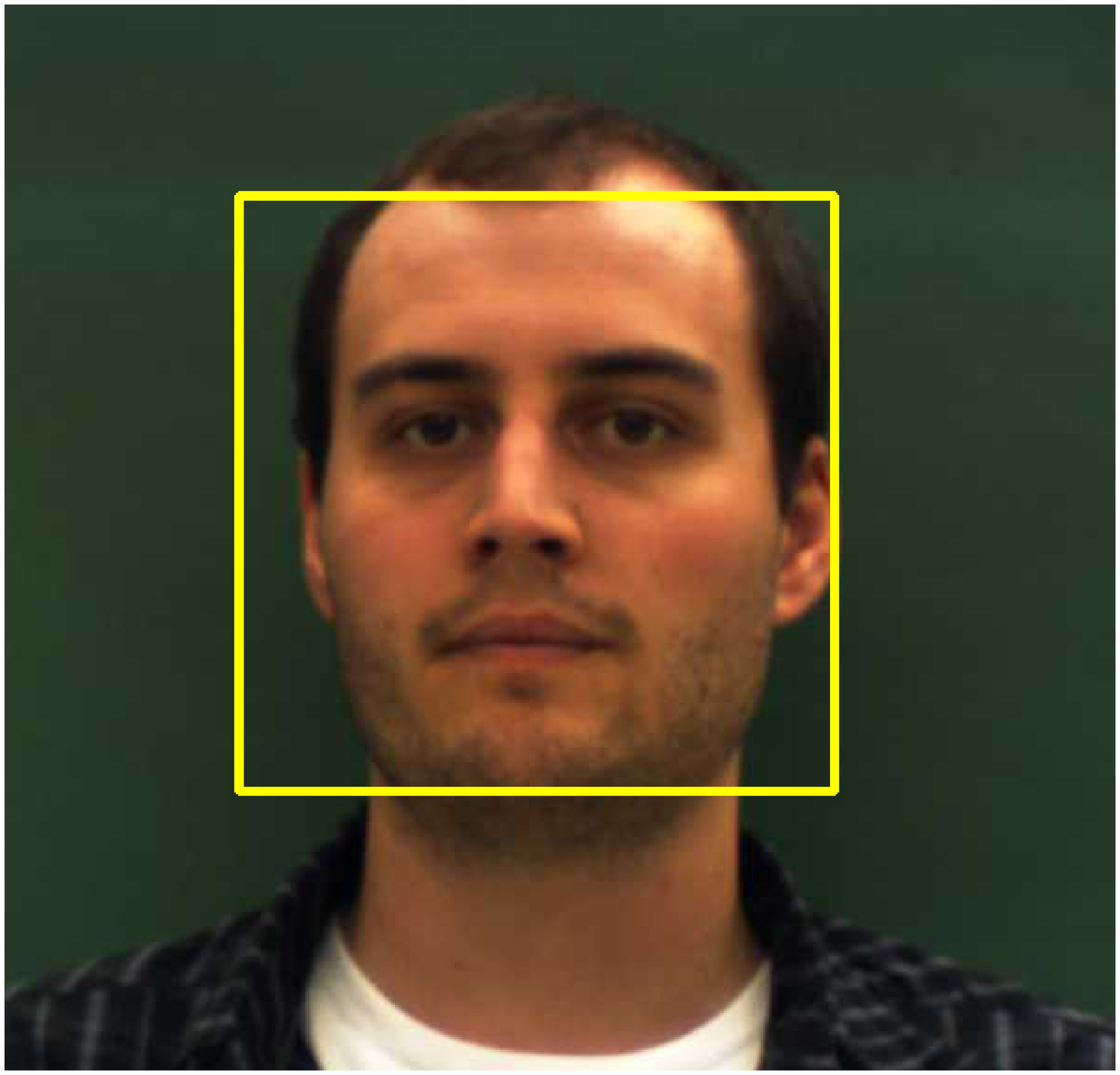} &
	\includegraphics[width=3.9cm]{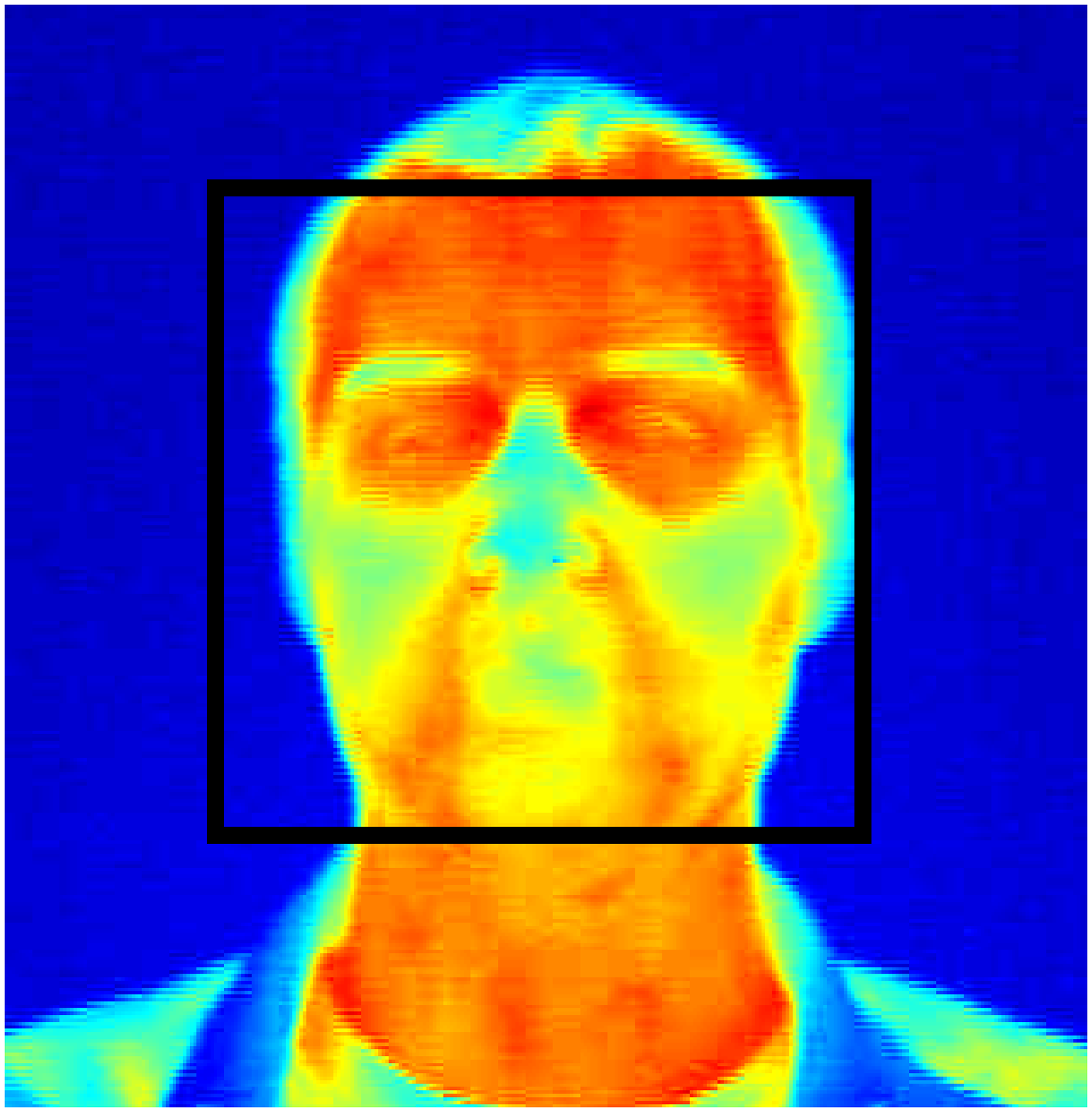} \\
	  \end{tabular}
	\caption{Automatic face detection within an RGB image, and location of region-of-interest superimposed on the IR image (JET colormap).  }
	\label{f.t2}
	\end{center}
	
\end{figure}

Figure \ref{f.t1} displays a set of 3 IR and RGB images where pose estimation is used to exclude rotated face (left, middle, right).  Since figure \ref{f.t1} (b) is a frontal view, additional information is extracted such as temperature and blood flow rate.  In addition, a rectangular region-of-interest is selected manually for further calculations showing a temperature of 33.727$^\circ$C and blood flowrate of 39.6536 $ml/100g tissue\cdot min$.  Figure \ref{f.t2} is an expanded view of a frontal frame where automatic face detection is applied and a color-map is applied to the IR image.  The measured room temperature is ~21.8$^\circ$C.

\begin{figure}[!ht]
\hspace{5mm}
  \includegraphics[width=8.2cm]{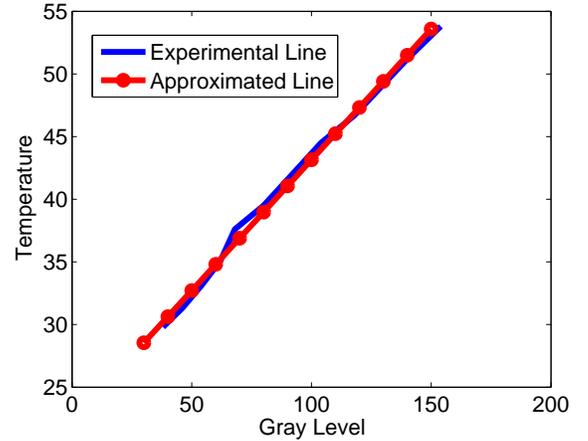} 
\caption{Temperature measured in Celsius degrees for equation estimation.}
	\label{f.temp}
\end{figure}

\section{Conclusions}

This work contributes towards development of a new generation of  access control systems that utilize biometrics for both authentication and situational awareness. In such systems,  biometrics such as  RGB-D, as well as Infrared facial images are used to  support the decision-making process. This support is facilitated by transferring the  data from biometric-sensing devices into an acceptable semantic form that supports an access control personnel in dialogue with a customer.

In our approach, the head pose estimation algorithm  filters out the unwanted head positions, and only performs facial recognition on the frontal view images. Instead of creating a system such as 3D modeling that accounts for rotation of the head, the approach proposed in this paper utilizes only frontal facial images. The latter are extracted from the video captured while a traveler is in the process of using the kiosk that reads the e-passport rather than forcing the traveler to  a special photoshoot location. For that purpose, we developed a face recognizer that combines the convenience of on-line video capturing and the seamless  frontal facial view detection. 

The presented experiments show that using depth data for the frontal view selection leads to a 35\% increase in face detection rate, as well as up to a 9\% increase in face recognition accuracy.  In addition, the frontal view selection also reduces the total processing time for the entire face recognition procedure by lowering the amount of frames required to run the face detection and recognition algorithms.  The overall reduction in processing time is approximately 32\%.  The frontal view selection pre-processing based on depth data leads to a reduction in time complexity and an improvement in face detection and recognition rates.  

In the context of access control systems, remote temperature estimation based on the IR facial images is aimed at assisting the personnel of the automated access control systems in their task of monitoring customers, in particular, in determining possible fever in the subjects.  Fusion of IR data with other facial biometrics is a source of data for complex intelligent access control systems, providing situational awareness and risk management.

The other future direction is to utilize depth information in order to create a 3D model using both 2D and depth information to enhance face recognition, or to perform fusion of such data at certain levels of recognition process.


\section*{Acknowledgment}
This project was implemented in the Biometrics Technology Laboratory at the University of Calgary, and partially supported by the Natural Sciences and Engineering Research Council of Canada (NSERC). K. Lai and S. Samoil acknowledge the Queen Elizabeth (II) Scholarship. 

{\small
\bibliographystyle{IEEEtran}
\bibliography{depth}

}
\end{document}